\documentclass[letterpaper, 10pt, conference]{ieeeconf}  
\pdfoutput=1
\IEEEoverridecommandlockouts                              

\usepackage{amsthm}
\usepackage{times}
\usepackage{multicol}
\usepackage[bookmarks=true]{hyperref}
\usepackage{xcolor}
\usepackage{hyperref}
\usepackage{amsmath, amssymb}
\usepackage{amsfonts}
\usepackage{graphicx}
\usepackage{siunitx}
\usepackage{standalone}
\usepackage{booktabs}
\usepackage[ruled,vlined,linesnumbered,noend]{algorithm2e}
\usepackage{mdframed}
\usepackage{fancyvrb,multirow}
\usepackage{soul}
\usepackage{dsfont,mathabx}
\usepackage{array, booktabs}
\usepackage{subcaption}
\usepackage{booktabs}
\usepackage{makecell}
\usepackage{tikz}
\usepackage{pgfplots}
\usepackage[labelfont=bf,format=plain]{caption}
\usepackage[para,online,flushleft]{threeparttable}

\theoremstyle{definition}

\newtheorem{definition}{Definition}
\newtheorem{remark}{Remark}

\title{\textbf{FlyKites: Human-centric Interactive Exploration\\
and Assistance under Limited Communication}}

\author{Yuyang Zhang$^*$, Zhuoli Tian$^*$, Jinsheng Wei and Meng Guo
  \thanks{The authors are with the College of Engineering,
    Peking University, Beijing 100871, China.
    This work was supported by the National Natural Science Foundation
    of China (NSFC) under grants 62203017, U2241214, T2121002.
    $^*$Equal Contribution.
    Contact: {\tt\small meng.guo@pku.edu.cn}.}
  }

\begin{document}
\maketitle
\thispagestyle{empty}
\pagestyle{empty}


\begin{abstract}
  Fleets of autonomous robots have been deployed for exploration 
  of unknown scenes for features of interest,
  e.g., subterranean exploration, reconnaissance, search and rescue missions.
  During exploration, the robots may encounter un-identified targets, 
  blocked passages, interactive objects, temporary failure, 
  or other unexpected events,
  all of which require consistent human assistance with reliable 
  communication for a time period.
  This however can be particularly challenging if 
  the communication among the robots is severely restricted
  to only close-range exchange via ad-hoc networks, 
  especially in extreme environments like caves and underground tunnels.
  This paper presents a novel human-centric interactive exploration
  and assistance framework called FlyKites,
  for multi-robot systems under limited communication.
  It consists of three interleaved components: 
  (I) the distributed exploration and intermittent communication 
  (called the ``spread mode"),
  where the robots collaboratively explore the environment and 
  exchange local data among the fleet and with the operator;
  (II) the simultaneous optimization of the relay topology, 
  the operator path, and the assignment of robots to relay roles
  (called the ``relay mode"),
  such that all requested assistance can be provided with minimum delay;
  (III) the human-in-the-loop online execution, 
  where the robots switch between different roles 
  and interact with the operator adaptively.
  Extensive human-in-the-loop simulations and hardware experiments are performed
  over numerous challenging scenes.
\end{abstract}

\section{Introduction}\label{sec:intro}
Exploration of unknown and hazardous scenes before allowing humans inside
is particularly suitable for robots.
For instances, fleets of UAVs and UGVs have been deployed 
to explore planetary caves in~\cite{klaesson2020planning, petravcek2021large};
search and rescue after earthquakes in~\cite{couceiro2017overview}.
Many collaborative exploration strategies have been
proposed along with the advances in autonomous navigation and perception,
e.g.,~\cite{colares2016next,hussein2014multi,yamauchi1997frontier,patil2023graph,
  zhou2023racer}.
They often assume an all-to-all communication
among the robots, i.e., instant map sharing among the robots
and with a static base station.
However, this could be impractical in many aforementioned scenes
where the communication facilities are unavailable or severely degraded.
Namely, the robots can only exchange information 
via ad-hoc networks in close proximity.
This imposes great challenges on the fleet coordination 
as communication and exploration are now closely dependent 
thus need to be jointly planned.


\begin{figure}[t]
  \centering
  \includegraphics[width=1.0\linewidth]{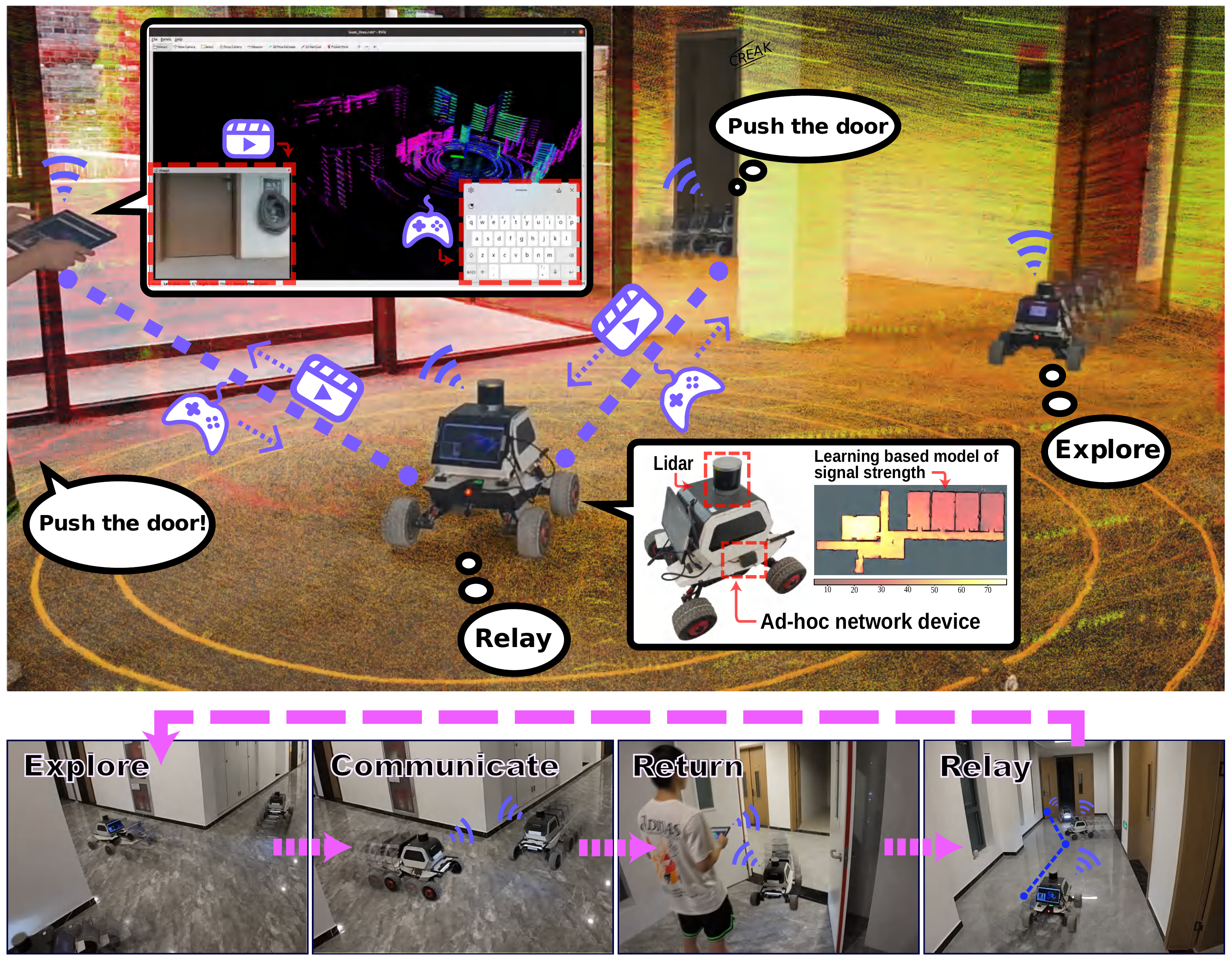}

  \caption{
  \textbf{Top}: the FlyKites system deployed for hardware experiments.
  One robot explores the workspace,
  while two robots in relay mode to transmit:
  (I) live video stream from the end robot to the operator,
  and (II) tele-operation from the operator to the end robot to open a door.
  Their communication is fully distributed via ad-hoc networks,
  of which the quality is predicted via a learned model.
  \textbf{Bottom}: Three UGVs transit between exploration, 
  intermittent communication, return to the operator, and communication relay. 
  }\label{fig:overall}
  \vspace{-6mm}
\end{figure}

Moreover, the robots often encounter situations where they need human
assistance during the exploration process~\cite{murphy2004rescue}.
For instance, the planned passage might be too tight for autonomous navigation,
which requires the operator to drive the robot manually~\cite{Podnar2006HumanTO};
the robot might encounter un-identified targets that need human inspection~\cite{cai2019inspect};
some objects might be interactive such as movable obstacles and doors to push open,
as shown in Fig.~\ref{fig:overall};
or the robots are simply stuck and need human intervention.
In these cases, a reliable communication between the requested robot and 
the operator is crucial for a period of time until the request is resolved~\cite{Marchukov2019chain}.
This however is particularly challenging to fulfill, 
if the robots and the operator can only communicate in close range 
in these extreme environments.

\subsection{Related Work}
Multi-robot exploration has been extensively studied in the 
literature~\cite{burgard2005coordinated, colares2016next, hussein2014multi,
patil2023graph, yamauchi1999decentralized, zhou2023racer}.
These work often assume that all robots can
exchange information via wireless communication
perfectly and instantly at all time.
This is often impractical or infeasible without infrastructures
for communication already installed.
As discussed in~\cite{esposito2006maintaining}, 
due to obstructions of obstacles in subterranean or indoor structures,
the inter-robot communication is severely limited in range and bandwidth.
To overcome this challenge,
many recent work can be found that combines on the planning for
of inter-robot communication and autonomous exploration.
The survey in~\cite{banfi2015communication} provides a quantitative
comparison of different communication-constrained exploration strategies,
including~\cite{pei2013connectivity, rooker2007multi}.
Different intermittent communication strategies have been proposed
w.r.t. different performance metrics~\cite{cesare2015multi,gao2022meeting,
  guo2018multirobot, marchukov2019fast,schack2024sound,vaquero2018approach}.
On the other hand, all-time fully-connected networks are imposed
in~\cite{rooker2007multi,zavlanos2011graph} by collaborative motion planning,
and further in~\cite{marcotte2020optimizing, pei2013connectivity} 
by placing front and relay nodes.
However, most these works focuse one specific mode of communication,
while neglecting the online interaction with human operators.

Indeed, the human operator plays {an indispensable role}
for the operation of robotic fleets,
despite of their autonomous capability.
In many aforementioned scenarios,
not only the operator should be aware of the system status but also
directly assist certain procedures whenever necessary.
Almost all aforementioned work neglects this aspect and
assume a static base station for visualization,
see~\cite{gao2022meeting, klaesson2020planning, 
marchukov2019fast, pei2010coordinated, saboia2022achord,vaquero2018approach},
yielding a rather uniform behavior of the fleet.
An augmented-reality (AR) device is proposed in~\cite{reardon2019communicating}
to facilitate the human-fleet real-time interaction,
which however relies on high-bandwidth communication.
The aforementioned bilateral and online interactions remains largely 
unexplored in the literature, i.e., how different 
communication modes can enable these online interactions.

\subsection{Our Method}
To tackle these challenges, this work proposes a novel framework FlyKites,
for the online interaction and assistance between the operator
and the robotic fleet in unknown and communication-constrained environments.
In particular, a generic collaborative exploration scheme is proposed for 
multi-robot systems with intermittent communication 
(called the ``spread mode"),
where the inter-robot communication and frontier-based exploration
are jointly planned between pairs of robots.
Then, to fulfill the robot requests for human assistance during exploration, 
a hybrid optimization problem is formulated to simultaneously optimize 
the relay topology, the operator path and the associated assignment of 
robots to relay roles, such that a reliable chain of communication
can be established between the operator and the requested robot.
Thus, the requested assistance can be provided with minimum delay,
thus called the ``relay mode".
Finally, an online strategy for human-in-the-loop execution is proposed,
where the robots switch between the exploration, communication and relay modes,
according to the exploration progress and the online interactions.
Extensive human-in-the-loop simulations and hardware experiments are performed
over numerous challenging scenes.

Main contributions of this work are two-fold:
(I) a novel and generic framework for the online interaction and assistance 
between a dynamic operator and a robotic fleet in unknown and communication-constrained environments;
(II) two different modes of simultaneous exploration and communication, 
i.e., the spread mode and the relay mode, 
and the transition strategy between them.

\section{Problem Description}\label{sec:problem}

\subsection{Robots and Operator in Workspace}\label{subsec:ws}
Consider a 2D workspace~$\mathcal{A}\subset \mathbb{R}^2$,
of which its {map} including the boundary, freespace 
and obstacles are all unknown.
A team of robots denoted by~$\mathcal{N}\triangleq\{1,\cdots,N\}$ 
is deployed by an operator to explore the workspace.
Each robot~$i\in \mathcal{N}$ is capable of simultaneous localization 
and mapping (SLAM) with collision avoidance.
Denote by~$p_i(t)\in \mathcal{A}$ the 2D pose and $M_i(t)\subseteq \mathcal{A}$ 
the local map of robot~$i$ at time~$t>0$.
Similarly, the operator has a~2D pose~$p_{\texttt{h}}(t)$ 
and a local map~$M_\texttt{h}(t)$.
For brevity, denote by~$\mathcal{N}^+\triangleq \mathcal{N}\cup \{\texttt{h}\}$.
Moreover, the robots and the operator are equipped 
with a communication module to exchange data locally as follows.
\begin{definition}[Neighbors]\label{def:com}
Each robot~$i\in \mathcal{N}^+$ (or the operator) 
can communicate with another neighboring robot~$j\in \mathcal{N}^+$ 
(or the operator),
if the communication quality between them is above a threshold, 
i.e., $\texttt{Com}_{ij}(p_i,p_j,M_i)>\underline{c}$.
$\underline{c}$ represents the minimum signal strength required 
for successful communication in a specific environment, 
which can be determined through experimental measurements.
\hfill $\blacksquare$
\end{definition}

The neighbors of robot~$i\in \mathcal{N}$ at time~$t>0$ is denoted 
by $\mathcal{N}_i(t)\subseteq \mathcal{N}^+$,
which is symmetric and time-varying.
Thus, the behavior of each robot~$i\in \mathcal{N}$ is determined
by its timed sequence of navigation and communication events, i.e.,
\begin{equation}\label{eq:behavior}
\Gamma_i\triangleq c^0_i\, \mathbf{p}^0_i\, c^1_i \, \mathbf{p}^1_i\,c^2_i \, \cdots,
\end{equation}
where~$c^m_i\triangleq (j,\, p_{ij}, \,[t_m,\,(t_m+T_{ij})])$ is 
the communication event with robot~$j\in \mathcal{N}_i(t)$ 
at location~$p_{ij}\in \mathcal{A}$ during the time interval~$[t_m,\,(t_m+T_{ij})]$;
the navigation path~$\mathbf{p}^m_i\subset \mathcal{A}$ contains the waypoints
between these communication events.
Similarly, 
the behavior of the operator is denoted by~$\Gamma_{\texttt{h}}$,
which however is not fully \emph{controllable}.

\subsection{Online Assistance via Human-robot Interaction}\label{subsec:robot-human}
During exploration, the robots may encounter situations that require human assistance,
e.g., to identify a target, to take over control inputs, 
to manipulate an interactive object,
or to resolve a temporary failure.
Thus, a set of \emph{unknown} assistance tasks for the fleet 
is defined as follows:
\begin{equation}\label{eq:assistance}
\Phi \triangleq \big\{\phi_1,\, \phi_2,\,\cdots,\phi_K \big\},
\end{equation}  
where~$\phi_k\triangleq (p_k,\, i_k,\,\rho_k,\,{T}_k)$ 
represents the $k$-th assistance task,
which includes the location~$p_k\in \mathcal{A}$,
the robot~$i_k\in \mathcal{N}$ requiring assistance,
the priority~$\rho_k>0$, and the minimum required duration~${T}_k>0$
for the task, 
$\forall k \in \mathcal{K} \triangleq \{1,\cdots, K\}$. 
Note that ${T}_k$ is  unknown and specified online by the operator.
\begin{definition}
[Condition for Assistance] \label{def:assistance}
An assistance task~$\phi_k\in \Phi$ is \emph{accomplished}
at time~$t_k>T_k$, under the following conditions:
(I)  robot~$i_k$ is at the location~$p_k$, 
with $p_{i_k}(t_k)=p_k$;
(II) there exists a {chain} of neighboring robots connecting 
robot~$i_k$ to the operator during the time 
period~$[(t_k-{T}_k),\, t_k]$, i.e.,
\begin{equation}\label{eq:line}
\boldsymbol{\xi}_k \triangleq 
i_{k}\, i^1_{k}\, i^2_{k}\,\cdots \, i^{L_k}_{k}\, \texttt{h},
\end{equation}
where~$i^\ell_{k}\in \mathcal{N}$ 
is the relay robot,~$\forall \ell \in \widehat{L}_k \triangleq \{1,\cdots,L_k\}$
with~$L_k$ relay robots in total;
and $i^{\ell+1}_k \in \mathcal{N}_{i^{\ell}_k}(t)$, 
$i^{1}_k \in \mathcal{N}_{i_k}(t)$, 
$\texttt{h} \in \mathcal{N}_{i^{L_k}_k}(t)$, 
$\forall \ell \in \widehat{L}_k$ and $\forall t \in [(t_k-T_k),\, t_k]$.
\hfill $\blacksquare$
\end{definition}

In other words, the conditions for assistance
require that a consistent and reliable communication chain
is established between the robot that needs assistance 
and the operator,
which is often assumed in related work~\cite{
colares2016next, hussein2014multi, patil2023graph, zhou2023racer}.
For brevity, denote by~$\widehat{\Gamma}(t) \triangleq 
\big{(}\{\Gamma_i(t)\}, \Gamma_{\texttt{h}}(t)\big{)}$ the joint behaviors 
of the robots and the operator by time~$t>0$.
Then, $\widehat{\Gamma}(t) \models \phi_k$ if the joint behaviors satisfy 
the above conditions for the assistance task~$\phi_k\in \Phi$.

\subsection{Problem Statement}\label{subsec:problem}
The overall problem is formalized as a constrained optimization over
the collaborative exploration and communication strategy over the fleet, i.e.,
\begin{subequations} \label{eq:problem}
  \begin{align}
    &\mathop{\mathbf{min}}\limits_{\{\widehat{\Gamma},\, \overline{T}\}}\;  \overline{T} \notag\\
    \textbf{s.t.}\quad & \mathcal{A} \subseteq M_\texttt{h}(\overline{T}); \label{subeq:terminal}\\
    & \widehat{\Gamma}(\overline{T}) \models \phi_k,\;\forall \phi_k \in \Phi; \label{subeq:assistance}
  \end{align}
\end{subequations}
where~$\overline{T}>0$ is the total time when 
the complete map~$\mathcal{A}$ is known to the operator by~\eqref{subeq:terminal},
and all online requests for assistance tasks are fulfilled by~\eqref{subeq:assistance}.

\section{Proposed Solution}\label{sec:solution}
The solution contains three main components:
(I) the distributed exploration and intermittent communication 
as the fundamental building block of ``spread mode";
(II) the simultaneous optimization of the relay topology, 
the operator path, and the assignment of robots to relay roles,
to fulfill the requested assistance, as the ``relay mode";
and (III) the human-in-the-loop online execution, 
where the robots switch between different modes 
and interact with the operator.

\subsection{Distributed Exploration and Intermittent Communication}
\label{subsec:exp-comm}

\subsubsection{Frontier-based Exploration}
The frontier-based method in~\cite{yamauchi1997frontier}
allows a robot to explore the environment
by repetitively reaching the frontiers on the boundaries between
the explored and unexplored areas of its local map.
Namely,
given the local map~$M_i(t)$ of robot~$i\in \mathcal{N}$ at time~$t>0$,
these boundaries can be identified via a Breadth-First-Search (BFS),
which are then clustered to a few frontiers by various
metrics~\cite{holz2010evaluating}.
Denote by~$\mathcal{F}_i(t)\triangleq \{f_k\}$ the set of frontiers,
where each frontier~$f_k\in \partial M_i$ is on the boundary~$\partial M_i$.
Thus, $\mathcal{F}_i(t)$ is empty if~$M_i(t)$ is fully explored.

\subsubsection{Intermittent Communication with Ring Topology}
\begin{figure}[t]
  \centering
  \includegraphics[width=0.8\linewidth]{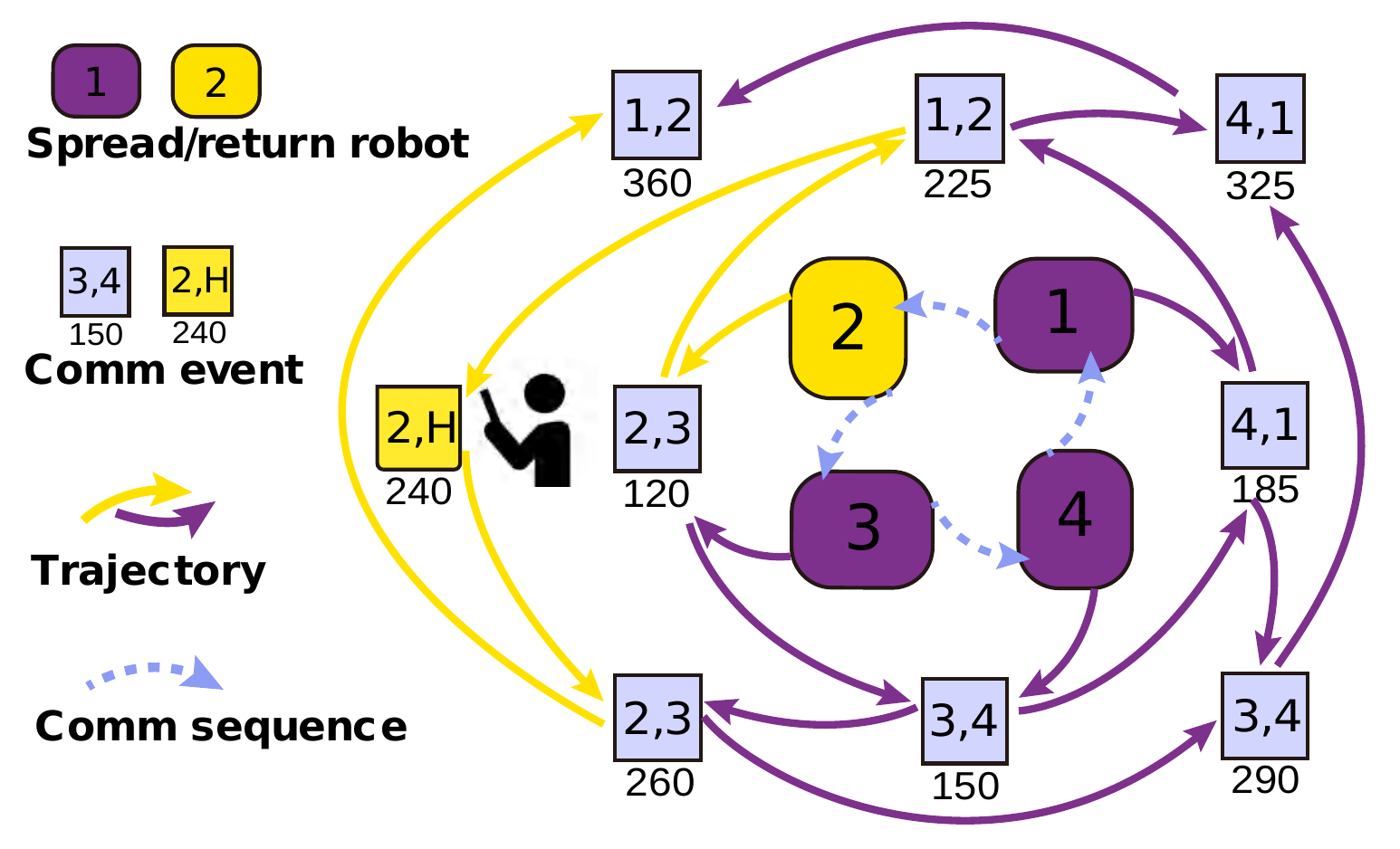}
  \caption{The intermittent communication protocol with the ring topology: 
  pairwise communication events (blue)
  and return events to the operator (yellow).
  }
  \label{fig:inter_comm_and_return}
  \vspace{-6mm}
\end{figure}

To ensure that the information can be propagated among the team 
and to the operator,
the robots are required to meet and communicate via 
{intermittent} communication during exploration.
As shown in Fig.~\ref{fig:inter_comm_and_return}, 
initially all robots are in close proximity but 
follows a fixed \emph{ring} topology.
Namely, each robot~$i$ only communicates with its predecessor~$i-1$ 
and successor~$i+1$, $\forall i=2,\cdots,(N-1)$;
robot~$N$ with robots~$1$ and $N-1$;
and robot~$1$ with robots~$N$ and $2$.
Without the need for assistance,
when robots~$i$ and $j$ communicate 
at a planned event~$c_{ij}\triangleq (t_{ij},p_{ij})$
with time~$t_{ij}>0$ and location~$p_{ij}\in M_i\cap M_j$,
they follow these steps:
(I) their local maps~$M_i$ and~$M_j$ are merged 
as the new local map, i.e., $M_{ij}\triangleq \texttt{merge}(M_i,M_j)$;
(II) the frontiers~$\mathcal{F}_{ij}$ within~$M_{ij}$ is
computed;
(III) the ``time and place'' for their \emph{next} communication event 
is determined by the constrained optimization problem:
\begin{equation}\label{eq:local-plan}
  \begin{split}
&(\boldsymbol{\tau}^+_i,\,\boldsymbol{\tau}^+_j) 
= \underset{(p^+_{ij},t^+_{ij})}{\textbf{argmax}}\; 
\Big{\{}\texttt{C2VRP}\big{(}(\boldsymbol{\tau}_i,\boldsymbol{\tau}_j),
p^+_{ij},\mathcal{F}_{ij}\big{)}\Big{\}}\\
&\textbf{s.t.}\qquad p^+_{ij}\in M_{ij}, \quad t^+_{ij}\leq t_{ij}
+\widehat{T}_{\texttt{h}}(t_{ij});
  \end{split}
\end{equation}
where~$(p^+_{ij},t^+_{ij})$ is the next communication event to be optimized;
$\boldsymbol{\tau}_i$ and~$\boldsymbol{\tau}_j$ are the current local plans of robots~$i$ and~$j$,
including the their confirmed meeting events with other robots;
function~$\texttt{C2VRP}(\cdot)$ stands for the constrained two-vehicle routing problem
to find the updated plans~$\boldsymbol{\tau}^+_i$ and~$\boldsymbol{\tau}^+_j$,
such that the robots can reach the maximum number of frontiers in~$\mathcal{F}_{ij}$,
while respecting the confirmed meeting events;
and~$p^+_{ij}$ should be chosen within the merged map,
while the time~$t^+_{ij}$ should be within a specified window~$\widehat{T}_{\texttt{h}}(t_{ij})$,
which is updated online as described in the sequel.
In other words, the robots should collaboratively determine the next communication event,
such that the exploration efficiency can be maximized.
The above problem can be solved either optimally via a mixed-integer programming (MIP) solver,
or approximately by solving two traveling salesman problem with time windows (TSPTW) 
in sequence;
(IV) after obtaining the updated plans~$(\boldsymbol{\tau}^+_i,\boldsymbol{\tau}^+_j)$, 
they depart and do not communicate until the confirmed communication event.

\begin{remark}\label{rm:spontaneous}
  During exploration, it often occurs that robot~$i$ meets with another robot~$k$
  on its way to the meeting event with robot~$j$,
  which is called a \emph{spontaneous} meeting event.
  In this case,
  they exchange their local data~$D_i$ and~$D_k$
  without coordinating the next meeting event,
  such that the communication topology~$\mathcal{G}$ can be maintained,
  instead of growing into a full graph. 
  \hfill $\blacksquare$
  \end{remark}
\subsubsection{Latency-bounded Return to Human Operator}
As previously discussed, it is imperative that the robots 
frequently return to the operator to provide timely updates on exploration progress, 
system status, and any potential assistance requests. 
The robot designated for this task is referred to as the \emph{messenger}.
To this end, a latency~$T_{\texttt{h}}$ is specified by the operator,
meaning that at least one robot should return to the operator every~$T_{\texttt{h}}$. 
This can be enforced by maintaining an estimated time stamp:
\begin{equation}\label{eq:time-stamp}
\widehat{T}_{\texttt{h}}(t)\triangleq T_{\texttt{h}}-t
+\underset{i\in \mathcal{N}}{\textbf{max}}\, \{t^i_{\texttt{h}}\},
\end{equation}
where~$t^i_{\texttt{h}}$ is the last time that robot~$i$ returned to the operator;
and~$\widehat{T}_{\texttt{h}}(t)$ is the estimated time left for the next return.
Thus, during the intermittent communication by~\eqref{eq:local-plan},
the time window~$\widehat{T}_{\texttt{h}}(t)$ is updated after 
exchanging~$\{t^i_{\texttt{h}}\}$.
Then, each robot checks if it can return immediately to the operator and 
meet the latency requirement,
\emph{after} they meet at the optimized event~$(p^+_{ij},t^+_{ij})$.
If not, it means that one of them should return to the operator before 
their next communication event.
Since the preceding robot~$i<j$ has the latest data, it would return to the operator 
after its next confirmed meeting event.
This results in another confirmed meeting event at~$p_{\texttt{h}}$ and estimated 
time of arrival~$t^i_{\texttt{h}}$,
which is appended to the local plan~$\boldsymbol{\tau}_i$.
Consequently, the same optimization problem~\eqref{eq:local-plan} is adopted given 
this updated plan to determine the next communication event for robot~$i$ and~$j$.
Similar analyses can be found in our previous work~\cite{tian2024ihero}.
\subsection{Hybrid Optimization for Assistance Tasks}\label{subsec:asssist}
During online exploration, the robots may encounter scenarios where they 
need human assistance, as defined in~\eqref{eq:assistance}.
For each assistance task~$\phi_k\in \Phi$, 
the critical condition for its accomplishment from Def.~\ref{def:assistance}
is that there exists a communication chain~$\boldsymbol{\xi}_k$ 
in~\eqref{eq:line} from robot~$i_k$ to the operator~$\texttt{h}$,
which is optimized in this part.

\subsubsection{Relay Topology Optimization}
To begin with, the relay topology of the chain communication is optimized.
namely the positions of the relay robots.
The collision-free shortest path from $p_k$ to $\texttt{h}$ in $M_{ij}$
from robot~$i_k$ to the operator~$\texttt{h}$ is computed as $\mathbf{p}^\star$. 
Then, starting from the first anchor~$p^{\ell}_k=\mathbf{p}^\star{[k_\ell]}$
where~$\ell=0$,
the next anchor~$p^{\ell+1}_k=\mathbf{p}^\star[k_{\ell+1}]$ is given by the largest
index~$k_{\ell+1}>k_\ell$ such that
$\texttt{Com}(p^{\ell}_k,\, \mathbf{p}^\star[k_{\ell+1}],\,{M}_{ij})>\underline{c}$ holds.
The above procedure terminates when~$k_{\ell}=|\mathbf{p}^\star|$,
namely the last anchor is the operator~$p^{L_k+1}_k=\texttt{h}$.
The resulting topology is denoted by~$\mathbf{p}_k$ for the $k$-th assistance task.

\begin{figure}[t]
    \centering
    \includegraphics[width=0.95\linewidth]{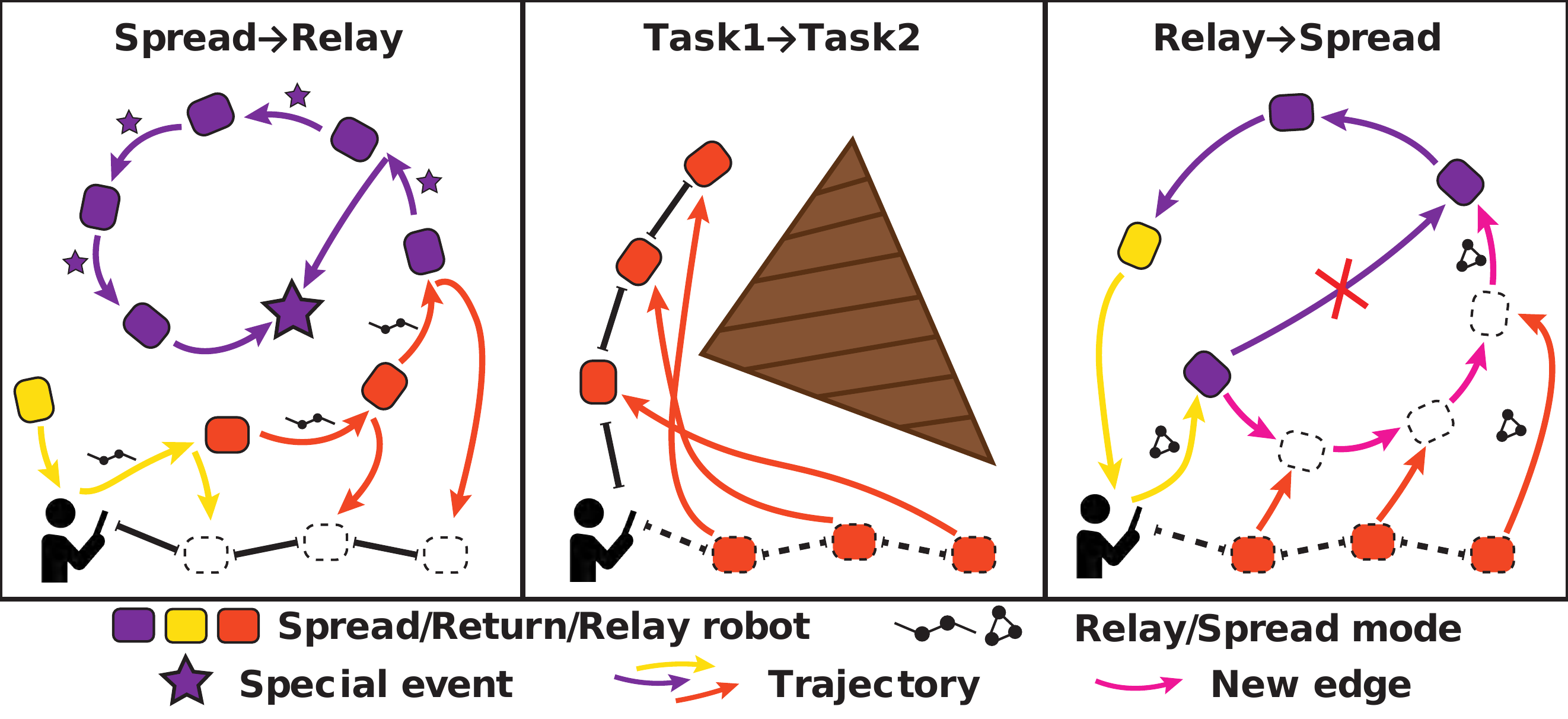}
    \caption{Mode transitions from the spread mode to the relay mode (\textbf{left}),
      directly between two relay modes (\textbf{middle}),
      or back to the spread mode (\textbf{right}).
    }
    \label{fig:s2r2s}
    \vspace{-6mm}
\end{figure}

\subsubsection{Assignment of Relay Roles}
Consider that~$L_k\leq N-1$, i.e., there are enough robots to form the relay chain.
To ensure that an assistance task can be accomplished as soon as possible, 
the succeeding $L_k$ robots of the messenger in the ring topology are assigned as relay robots,
for they are the earliest to finish their last meeting events. 
Denote by~$\mathcal{N}_k \subseteq \mathcal{N}$ the set of assigned robots.
The problem is to determine 
the optimal assignment~$\Pi_{k}:\mathbf{p}_{k} \rightarrow  \mathcal{N}_k$,
such that the time when all assigned robots reach the anchor points is minimized, i.e.,
\begin{equation}\label{eq:obj-assign}
\mathop{\mathbf{min}}\limits_{\Pi_{k}}
\Big\{\mathop{\mathbf{max}}\limits_{\ell \in \widehat{L}_k}\;
\big\{t_{i_k^{\ell}} + T_{\texttt{Nav}}\big(p_{i_k^{\ell}}(t_k^{\ell}),\, p^{\ell}_k\big)\big\}\Big\},
\end{equation}
where~$\Pi_{k}(p^{\ell}_k)\triangleq (i_k^{\ell},\,t_k^{\ell})$, $\forall \ell=1,\cdots,L_k$;
$i_k^{\ell}\in \mathcal{N}_k$ is the robot assigned to the anchor point~$p^{\ell}_k$;
$t_k^{\ell}$ is the time when robot~$i_k^{\ell}$ navigates from its
final meeting event~$p_{i_k^{\ell}}(t_k^{\ell})$ towards~$p^{\ell}_k$.
The above problem resembles the linear bottle-neck assignment problem (LBAP), 
where the objective is to minimize the maximum cost of an agent-task assignment
(i.e., not the summation).
It can be solved in a centralized manner by the messenger, 
since it has obtained the planned meeting events of all other robots 
due to the ring topology.

\subsubsection{Operator's movement}
The operator may relocate upon each messenger's return to:
(I) reduce relay robots when~$L_k>N-1$, 
by moving to an anchor point along topology~$\mathbf{p}_k$;
(II) shorten the messenger's return by moving in the main exploration direction, 
speeding exploration.
The messenger records and distributes the operator's location.
\subsection{Mode Transitions}\label{subsec:transit}
Given the assignments, some relay robots must transition 
between spread and relay modes. As shown in Fig.~\ref{fig:s2r2s}, 
this section details the bilateral transition process.

\begin{algorithm}[t]
\caption{Online Execution and Adaptation}
\label{alg:online}
	\LinesNumbered
        \SetKwInOut{Input}{Input}
        \SetKwInOut{Output}{Output}
\Input{$\Phi$.}
\Output{$\{\Gamma_i\}$, $M_\texttt{h}$.}
\While{$M_{\texttt{h}} \neq \mathcal{A}$}{
\tcc{\textbf{Spread Mode}}
  \For{neighbors~$(i,j)$}{
    Compute~$M_{ij}$ and~$\mathcal{F}_{ij}$\;
    Update~$\Gamma^+_i, \Gamma^+_j$ by~\eqref{eq:local-plan}\;
    \If{return event required by~\eqref{eq:time-stamp}}{
        Return to operator and update~$M_{\texttt{h}}$, $\Phi$\;
    }
  }
  \tcc{\textbf{Relay Mode}}
  \If{assistance~$\phi_k\in \Phi$ known at the operator}{
    Compute assignment~$\Pi_k, \mathcal{I}, \mathcal{J}$ by~\eqref{eq:obj-assign}\;
    Robots~$\mathcal{I}$ transit to the relay mode\;
    \If{$\phi_k$ is completed}{
      All robots transit back to the spread mode\;
    }
  }
}
\end{algorithm}

\subsubsection{Transition From Spread Mode to Relay Mode}
Once the relay topology and role assignments are set, 
the fleet transitions to relay mode,
following the same intermittent communication protocol. 
The key challenge is ensuring the remaining robots 
maintain the ring topology and continue exploration in spread mode.
Assume that robot~$i^{\star}\in \mathcal{N}$ returns to the operator; 
$\mathcal{I}\triangleq i^{\star} i^1_k i^2_k \cdots i^{L_k}_k$ are the 
relay robots;  
and $\mathcal{J} \triangleq j^{1}_k j^{2}_k \cdots j^{\overline{L}_k}_k$ 
are the remaining robots in the spread mode, 
where $\overline{L}_k\triangleq N-L_k-1$. 
Then, the transition phase consists of three stages:
(I)
Each relay robot $i^{\ell}_k\in \mathcal{I}$ propagates the relay decisions
to its succeeding robot $i^{\ell+1}_k$, 
and then directly navigates to its assigned anchor point; 
(II) Robot~$j^{1}_k$ changes its preceding robot from~$i^{L_k}_k$ 
to $j^{\overline{L}_k}_k$, 
and updates their next meeting event~$c(j^{1}_k,j^{\overline{L}_k}_k)$ 
by~\eqref{eq:local-plan}; 
(III) This updated event is propagated among $\mathcal{J}$ 
and finally to robot~$j^{\overline{L}_k}_k$,
which changes its succeeding robot to~$j^{1}_k$. 
Through this procedure, 
the robots in~$\mathcal{I}$ transit to the relay mode 
for the assistance task, 
while the robots in~$\mathcal{J}$ remain in the spread mode 
for exploration.

\subsubsection{Transition From Relay Mode to Spread Mode}
Upon completing the assistance task, relay robots transition back to spread mode.
This is triggered when the messenger returns with updated data at~$t_{\texttt{r}}>0$,
obtains the remaining task time~$t_{\texttt{w}}>0$ from the operator, 
and checks the condition:
\begin{equation}\label{eq:relay-to-spread}
t_{\texttt{r}}+t_{\texttt{w}}+\mathop{\mathbf{max}}\limits_{\ell,j} 
\big\{ T_{\texttt{Nav}}(p^{\ell}_k,p^{\star}_j)\big\} \leq  
\mathop{\mathbf{max}}\limits_{j} \{t^{\star}_j\}, 
\end{equation}    
where~$t^{\star}_j$ and $p^{\star}_j$ denote 
the \emph{last} planned meeting time and location 
by the robots in the spread mode.
The above condition ensures that any relay robot 
can reach at least one planned communication event.
Afterwards, the transition from relay mode to spread mode proceeds as follows:
(I) The earliest communication event~$c_{j^\star}$ that satisfies~\eqref{eq:relay-to-spread}
is determined, with~$c_{j^\star+1}$ being the subsequent event in the spread mode. 
A collision-free path from~$p_{j^\star}$ to $p_{j^\star+1}$ is then computed as~$\mathbf{p}_{j^\star}$;
(II) Each relay robot~$i\in \mathcal{I}$
finds the nearest point $\widehat{p}_i \in  
\mathbf{p}_{j^\star}$ and computes its arrival time 
$\widehat{t}_{i}\triangleq t_{\texttt{r}}+t_{\texttt{w}}+T_{\texttt{Nav}}(p_i,\widehat{p}_i)$. 
Thus, the relay robots are re-ordered by their arrival time to~$\mathbf{p}_{j^\star}$ as $i_0 i_1 \cdots i_{L_k}$;
(III) Each relay robot~$i_\ell$ communicates with its predecessor~$i_{\ell-1}$ 
at event~$c({i_{\ell-1},i_\ell})$ 
and with its successor~$i_{\ell+1}$ at event~$c({i_{\ell},i_{\ell+1}})$,
$\forall \ell = 1,\cdots,{L_{k-1}}$.
Moreover, robot~$i_0$ communicates with its predecessor $j^\star$ at event~$c({j^\star},i_{0})$ 
and with its successor~$i_{1}$ at event~$c({i_{0},i_{1}})$;
robot~$i_{L_k}$ communicates with its predecessor $i_{L_{k-1}}$ at event~$c({i_{L_{k-1}},i_{L_{k}}})$ 
and with its successor~$j^\star +1$ at event~$c({i_{L_{k}},{j^\star +1}})$.
In this way, robot~$i_{L_k}$ effectively replaces the original predecessor of robot~$j^\star + 1$.
The communication events are then defined as follows: 
$c({i_{\ell-1},i_\ell})=(\widehat{t}_{i_\ell},\widehat{p}_{i_\ell})$,
$c({j^\star},i_{0})=(\widehat{t}_{i_0},\widehat{p}_{i_0})$,
and $c({i_{L_{k}},{j^\star +1}})=c_{j^\star +1}$.
Through this procedure, the planned communication 
events are updated, by which the robots in~$\mathcal{I}$ can transit 
back to the spread mode.
\subsection{Online Execution and Adaptation}\label{subsec:online}



As summarized in Alg.\ref{alg:online}, 
the execution alternates between spread mode, return events, and relay mode. 
Initially, all robots follow the spread mode for exploration and communication from Sec.\ref{subsec:exp-comm},
which governs inter-robot communication and return events. 
Upon receiving assistance requests, 
relay topology and roles are optimized via the hybrid method in Sec.\ref{subsec:asssist}, 
and assigned relay robots transition according to Sec.\ref{subsec:transit}. 
After completing a task, they return to spread mode unless another task is queued, 
in which case they navigate directly to the next without reverting. 
The operator prioritizes tasks based on~${\rho_k}$ when multiple requests arrive.
\begin{figure}[t]
  \centering
  \includegraphics[width=0.97\linewidth]{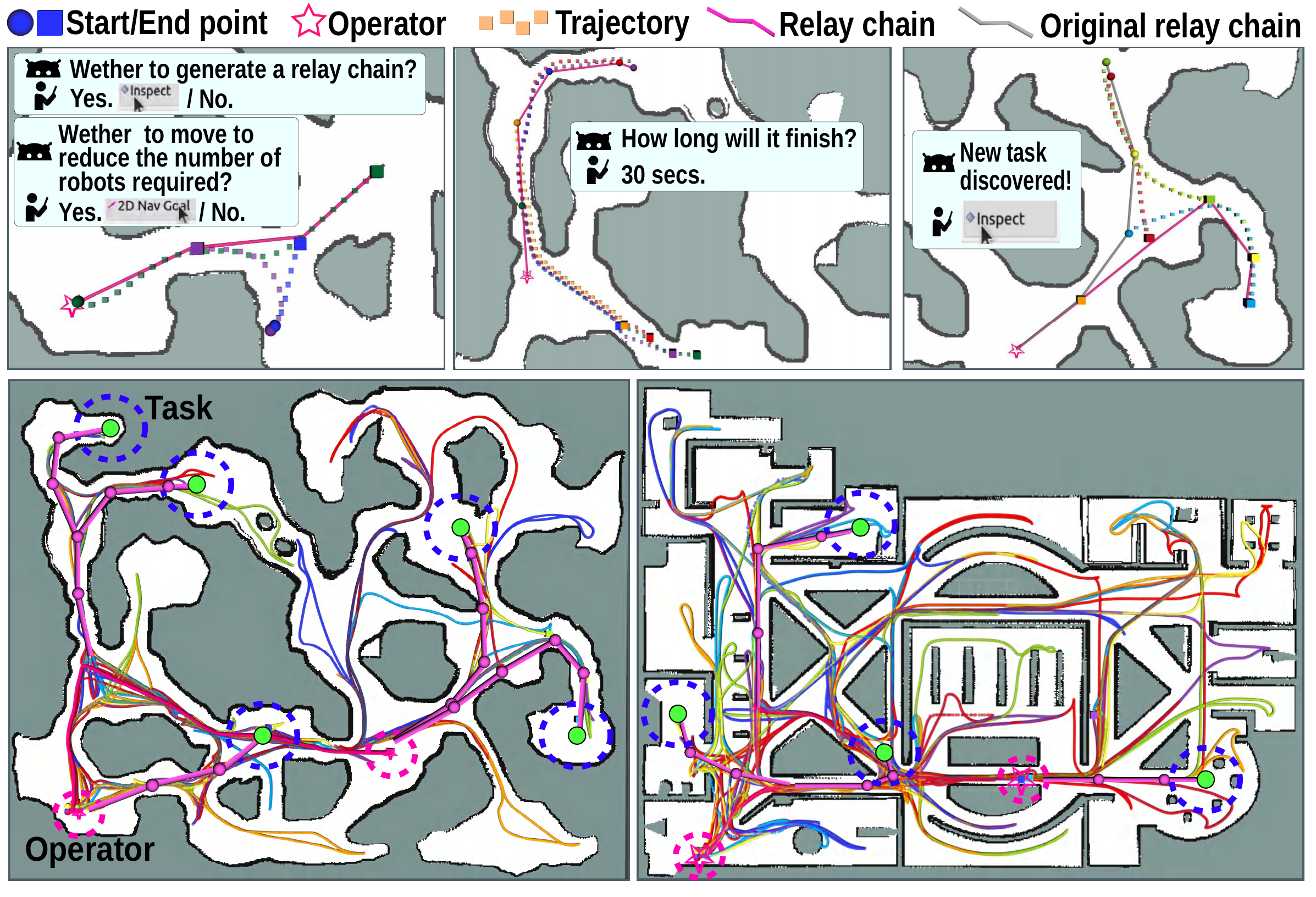}
  \caption{Representative simulation results.
    Online interactions between the operator and the fleet during
    three requests (\textbf{Top}).
    Fully-explored cave and emporium
    with robots and operator trajectories,
    where green dots indicate assistance tasks
    and purple dots indicate relay robots.(\textbf{Bottom}).
    \vspace{-0.1in}
  }\label{fig:sim2}
  \vspace{-4mm}
\end{figure}
\section{Numerical Experiments} \label{sec:experiments}
For further validation,
numerical simulations and hardware experiments are presented in this section.
The proposed method is implemented in \texttt{Python3}
within the framework of \texttt{ROS},
and tested on a computer with an Intel Core i7-13700KF CPU.
Simulation and experiment videos can be found in the supplementary files.

\subsection{System Description}\label{subsec:description}
The robotic fleet consists of~$8$ differential-driven UGVs,
which are  simulated in the \texttt{Stage} simulator and
visualized in the \texttt{Rviz} interface.
As shown in Fig.~\ref{fig:sim2}, two different workspaces are tested:
(I) a large subterranean cave of size $65m\times 57m$ with numerous tunnels;
(II) a large emporium of size $60m\times 50m$ with many connected rooms.
The occupancy grid map~\cite{moravec1985high} is adopted with a resolution of~$0.1m$
and a sensor range of~$10m$, generated via the $\texttt{gmapping}$ SLAM package.
Each robot navigates using the navigation stack \texttt{move\_base},
with a maximum linear velocity of~$0.6m/s$ and angular velocity of~$1.5rad/s$.
The operator could move with a velocity of~$0.3m/s$.
The robots and the operator start initially from the bottom-left corner of the map.

Moreover, two robots can only communicate if
they have a line of sight (LOS) and are within a range of~$5m$,
the same between robots and the operator.
The periodic update by~$T_{\texttt{h}}$ to the operator is set to~$60s$.
As shown in Fig.~\ref{fig:sim2}, five (Cave) or four (Emporium) assistance tasks are initially unknown,
and discovered by robots gradually.
Lastly, the operator interacts with the robotic fleet through the terminal
and a customized GUI in~\texttt{Rviz}.

\subsection{Simulation Results}\label{subsec:results}

\subsubsection{Subterranean Cave}
\begin{figure}[t]
  \centering
  \includegraphics[width=0.97\linewidth]{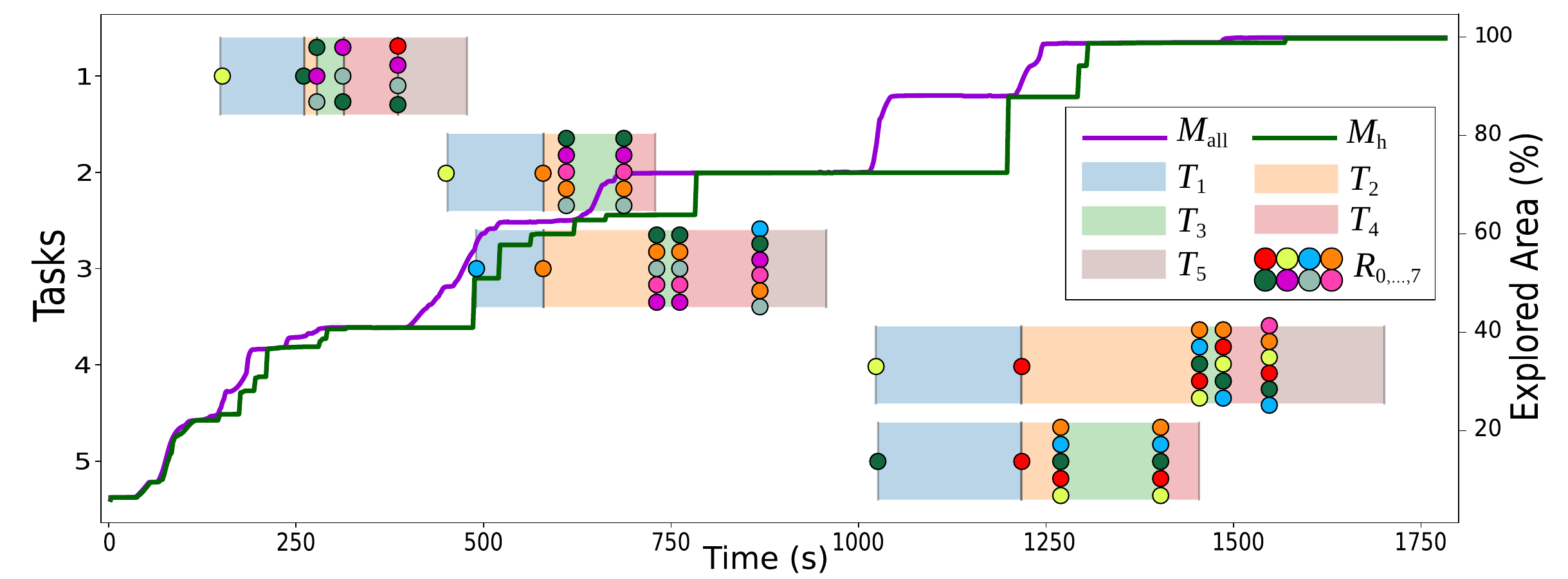}
  \caption{
    Progress of exploration and task fulfillment by~$8$ robots
    in the cave environment.
    Participants of five stages of each assistance task are indicated
    by filled dots;
    the map of the operator~$M_{\texttt{h}}$
    follows the union of robot maps~$M_{\texttt{all}}$.
  }\label{fig:task-stage}
  \vspace{-5mm}
\end{figure}

As shown in Fig.~\ref{fig:task-stage},
the fleet fully explores the~$1903$ square meters
within~$1498s$, while the operator receives the full map in $1571s$. 
Moreover, during exploration,
\textbf{five} assistance tasks are received and completed,
the last of which at~$1548s$.
Assistance task stages are recorded as in Fig.~\ref{fig:task-stage}:
$T_1$: time from discovery to operator reception;
$T_2$: operator interaction;
$T_3$: transition from spread mode to relay mode;
$T_4$: task execution via chain ($T_k$);
$T_5$: transition from relay mode to spread mode.
Fig.~\ref{fig:sim2} shows the $T_3$ of Task~$1$,
the $T_5$ of Task~$3$,
the directly transition in relay mode from Task~$5$ to~$4$.
Note that the participants of each assistance task are different
and several tasks are fulfilled in parallel.
It is worth emphasizing that the average transition time
(including from and back to spread mode)
for all tasks is around~$130.8s$.
Lastly, via the proposed scheme of direct transition
among assistance tasks,
robots~$3$-$7$ transit from Task~$2$ to~$3$ within only~$32s$, 
which is significantly shorter than the average duration of $64s$. 
Similar phenomenon occurs during the concurrent execution of Tasks~$4$ and~$5$.

\subsubsection{Emporium}
For the emporium environment, 
the structure of map is more complicated for exploring,
as shown in Fig.~\ref{fig:sim2},
in total \textbf{four} assistance tasks are imposed and completed in~$1536s$.
Tasks~$1$ and~$2$ are completed via direct transition at~$474s$, 
with a transition time of $40s$.
Tasks~$3$ and~$4$ are completed at $1020s$ and $1500s$, respectively.
In tasks~$3$ and~$4$, the operator choose to move along the relay topology,
reducing the number of relay robots by $2$ and $5$, respectively,
allowing more robots to continue exploring the map. 
Finally, the fleet fully explores the~$1737$ square meters within~$1591s$,
and the complete map is obtained by the operator at~$1620s$.

\subsection{Comparison}\label{subsec:comparison}
The proposed framework \textbf{FlyKites}
is compared against four baselines: 
(I) \textbf{SEP} separates exploration and task execution,
handling all requests post-exploration; 
(II) \textbf{OP-DM}, where operator directly moves to the requests without a chain;
(III) \textbf{OP-STA}, where the operator stays static; 
and (IV) \textbf{NO-DT}, 
where the direct transition among tasks is not allowed.
The compared metrics are the completion rate and time of all tasks,
and the average and maximum latency of all tasks, 
where task latency refers to the duration of time between observing a task and finishing it.

Table~\ref{table:table-data} summarizes the results of multiple simulation runs.
It can be seen that \textbf{FlyKites} achieves the best performance 
over all baselines for all metrics.
Although \textbf{SEP} completes tasks with only slightly longer time than \textbf{FlyKites}, 
it suffers from much larger task latency due to separation of exploration and task execution.
Without the proposed chain of communication,
the completion time of \textbf{OP-DM} is almost twice that of
ours ($3105s$ vs. $1571s$).
Note that~\textbf{NO-DT} has also a long time ($2792s$), 
as significant time is spent on mode transition
when there are multiple tasks.
Lastly, if operator stays static as in \textbf{OP-STA},
only~$60\%$ tasks can be handled 
due to the inadequate number of robots to
form a chain for Tasks~$4$ and~$5$.
In contrast, 
our method requires the minimal duration to accomplish all tasks 
with the minimum latency.

\begin{table}[t]
  \begin{center}
    \caption{Comparison with Baselines.}\label{table:table-data}
    \vspace{-0.05in}   
    \setlength{\tabcolsep}{0.8\tabcolsep}   
    \centering
    \begin{tabular}{c c c c c}
       \toprule[1pt]
      \midrule
      \textbf{Method} & 
      \textbf{\makecell{Task Comp-\\letion [\%]}} 
      & \textbf{\makecell{Finish\\ Time [s]}} 
      & \textbf{\makecell{Avg. Task \\Latency [s]}}
      & \textbf{\makecell{Max. Task \\Latency [s]}}\\[.1cm]
      \cmidrule{1-5}
      \textbf{FlyKites} & \textbf{100} & \textbf{1571}  & \textbf{303.6} & \textbf{465.0}\\[.1cm]
      SEP         & 100          & 1647        & 894.0        & 1055.0 \\[.1cm]
      OP-DM         & 100          & 3105        & 423.6       & 743.0 \\[.1cm]
      OP-STA         & 60          & \textbackslash        & $\infty$       & $\infty$ \\[.1cm]
      NO-DT         & 100          & 2792        & 425.6       & 676.0 \\[.1cm]
      \midrule
      \bottomrule[1pt]
    \end{tabular}
  \end{center}
  \vspace{-5mm}
  \end{table}


\subsection{Hardware Experiments}\label{subsec:hardware}
\begin{figure}[t]
  \centering
  \includegraphics[width=0.96\linewidth]{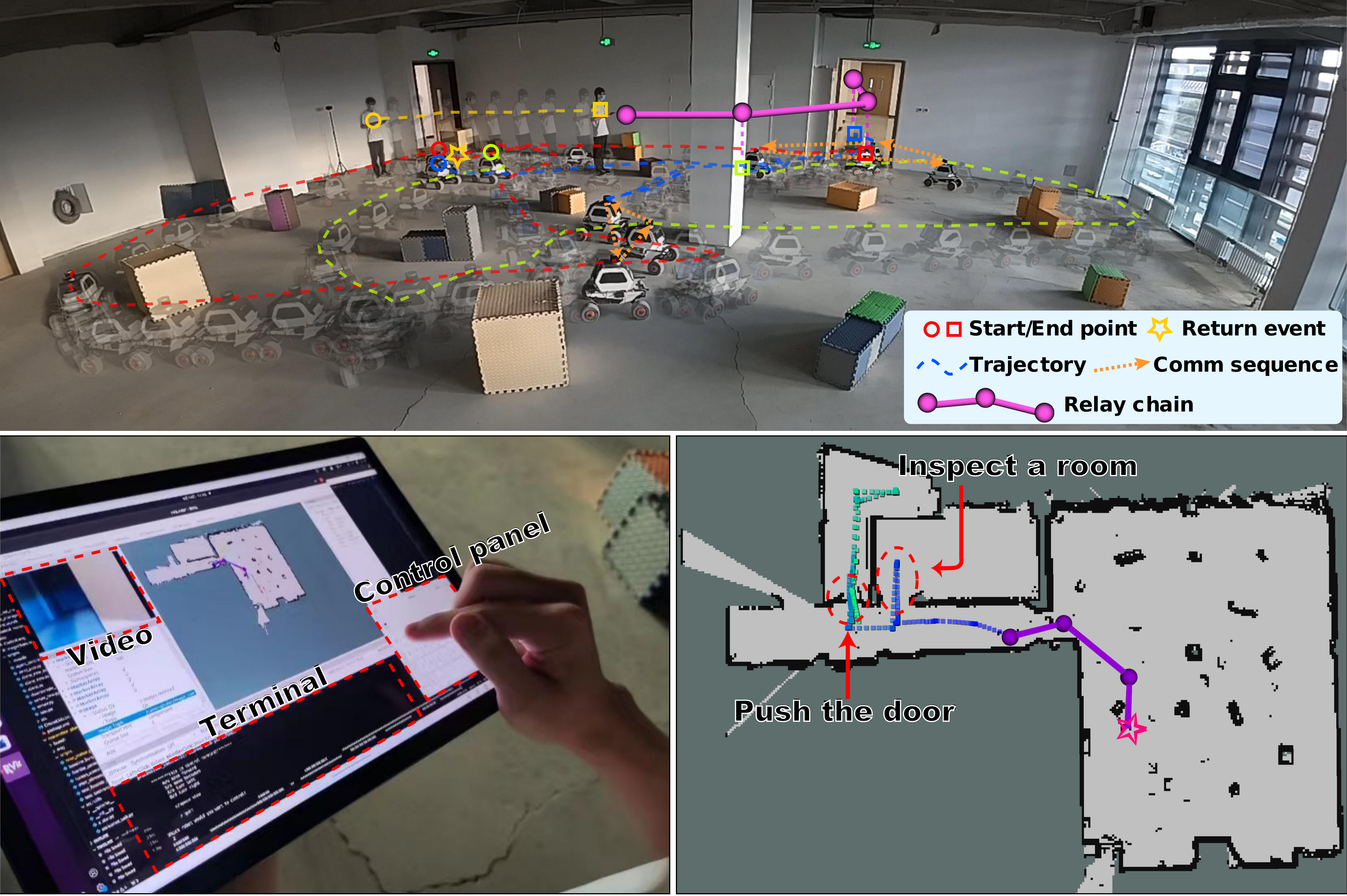}
  \caption{Hardware experiments where the operator interacts
    online with the fleet via a tablet and ad-hoc communication network.
    A communication chain is formed for the second task
    via three UGVs for video stream and tele-operation.}
  \label{fig:hardware}
  \vspace{-5mm}
\end{figure}

As shown in Fig.~\ref{fig:overall} and~\ref{fig:hardware},
an operator deploys $3$ ground robots to explore an office environment 
of size $35m\times 20m$, and interacts with the fleet with a tablet. 
Each robot and the operator are equipped with an ad-hoc network device 
for close-range communication (AP-DLINK1402A).
The office is fully explored at~$107s$, 
after which two robots explore the corridor
and the third one returns to operator. 
A communication chain of three robots is then formed at~$294s$, 
for which the operator moves along the chain
to reduce the number of relays from~$4$ to~$3$.
Then, the operator controls the end robot to 
(I) inspect a room; and (II) push the door of another room
to fully explore the environment.
Finally, the two requests are accomplished at~$523s$.
It is worth noting that via the live video stream from the end robot,
the operator can control it smoothly and safely \textbf{without} 
{direct line of sight}. 
\section{Conclusion} \label{sec:conclusion}
This work proposes a novel and generic framework (FlyKites)
for the online interaction and assistance 
between a dynamic operator and a robotic fleet
in unknown and communication-constrained environments.
Future work will focus on the integration with other collaborative tasks
besides exploration, e.g., search and rescue.


\newpage
\bibliographystyle{plain}
\bibliography{references}

\begin{thebibliography}{10}

\bibitem{banfi2015communication}
Jacopo Banfi, Alberto~Quattrini Li, Nicola Basilico, and Francesco Amigoni.
\newblock Communication-constrained multirobot exploration: Short taxonomy and
  comparative results.
\newblock In {\em Proceedings of the IROS workshop on on-line decision-making
  in multi-robot coordination (DEMUR2015)}, pages 1--8, 2015.

\bibitem{burgard2005coordinated}
Wolfram Burgard, Mark Moors, Cyrill Stachniss, and Frank~E Schneider.
\newblock Coordinated multi-robot exploration.
\newblock {\em IEEE Transactions on robotics}, 21(3):376--386, 2005.

\bibitem{cai2019inspect}
Hong Cai and Yasamin Mostofi.
\newblock Human-robot collaborative site inspection under resource constraints.
\newblock {\em IEEE Transactions on Robotics}, 35(1):200--215, 2019.

\bibitem{cesare2015multi}
Kyle Cesare, Ryan Skeele, Soo-Hyun Yoo, Yawei Zhang, and Geoffrey Hollinger.
\newblock Multi-uav exploration with limited communication and battery.
\newblock In {\em IEEE international conference on robotics and automation
  (ICRA)}, pages 2230--2235, 2015.

\bibitem{colares2016next}
Rafael~Gon{\c{c}}alves Colares and Luiz Chaimowicz.
\newblock The next frontier: Combining information gain and distance cost for
  decentralized multi-robot exploration.
\newblock In {\em Annual ACM Symposium on Applied Computing}, pages 268--274,
  2016.

\bibitem{couceiro2017overview}
Micael~Santos Couceiro.
\newblock An overview of swarm robotics for search and rescue applications.
\newblock {\em Artificial Intelligence: Concepts, Methodologies, Tools, and
  Applications}, pages 1522--1561, 2017.

\bibitem{esposito2006maintaining}
Joel~M Esposito and Thomas~W Dunbar.
\newblock Maintaining wireless connectivity constraints for swarms in the
  presence of obstacles.
\newblock In {\em IEEE International Conference on Robotics and Automation},
  pages 946--951, 2006.

\bibitem{gao2022meeting}
Yuman Gao, Yingjian Wang, Xingguang Zhong, Tiankai Yang, Mingyang Wang,
  Zhixiong Xu, Yongchao Wang, Yi~Lin, Chao Xu, and Fei Gao.
\newblock Meeting-merging-mission: A multi-robot coordinate framework for
  large-scale communication-limited exploration.
\newblock In {\em IEEE/RSJ International Conference on Intelligent Robots and
  Systems (IROS)}, pages 13700--13707, 2022.

\bibitem{guo2018multirobot}
Meng Guo and Michael~M Zavlanos.
\newblock Multirobot data gathering under buffer constraints and intermittent
  communication.
\newblock {\em IEEE Transactions on Robotics}, 34(4):1082--1097, 2018.

\bibitem{holz2010evaluating}
Dirk Holz, Nicola Basilico, Francesco Amigoni, and Sven Behnke.
\newblock Evaluating the efficiency of frontier-based exploration strategies.
\newblock In {\em International Symposium on Robotics}, pages 1--8. VDE, 2010.

\bibitem{hussein2014multi}
Ahmed Hussein, Mohamed Adel, Mohamed Bakr, Omar~M Shehata, and Alaa Khamis.
\newblock Multi-robot task allocation for search and rescue missions.
\newblock In {\em Journal of Physics: Conference Series}, volume 570, page
  052006, 2014.

\bibitem{klaesson2020planning}
Filip Klaesson, Petter Nilsson, Tiago~Stegun Vaquero, Scott Tepsuporn, Aaron~D
  Ames, and Richard~M Murray.
\newblock Planning and optimization for multi-robot planetary cave exploration
  under intermittent connectivity constraints.
\newblock 2020.

\bibitem{Marchukov2019chain}
Yaroslav Marchukov and Luis Montano.
\newblock Fast and scalable multi-robot deployment planning under connectivity
  constraints.
\newblock In {\em 2019 IEEE International Conference on Autonomous Robot
  Systems and Competitions (ICARSC)}, pages 1--7, 2019.

\bibitem{marchukov2019fast}
Yaroslav Marchukov and Luis Montano.
\newblock Fast and scalable multi-robot deployment planning under connectivity
  constraints.
\newblock In {\em IEEE International Conference on Autonomous Robot Systems and
  Competitions (ICARSC)}, pages 1--7, 2019.

\bibitem{marcotte2020optimizing}
Ryan~J Marcotte, Xipeng Wang, Dhanvin Mehta, and Edwin Olson.
\newblock Optimizing multi-robot communication under bandwidth constraints.
\newblock {\em Autonomous Robots}, 44(1):43--55, 2020.

\bibitem{moravec1985high}
Hans Moravec and Alberto Elfes.
\newblock High resolution maps from wide angle sonar.
\newblock In {\em IEEE International Conference on Robotics and Automation
  (ICRA)}, volume~2, pages 116--121, 1985.

\bibitem{murphy2004rescue}
R.R. Murphy.
\newblock Human-robot interaction in rescue robotics.
\newblock {\em IEEE Transactions on Systems, Man, and Cybernetics, Part C
  (Applications and Reviews)}, 34(2):138--153, 2004.

\bibitem{patil2023graph}
Indraneel Patil, Rachel Zheng, Charvi Gupta, Jaekyung Song, Narendar Sriram,
  and Katia Sycara.
\newblock Graph-based simultaneous coverage and exploration planning for fast
  multi-robot search.
\newblock {\em arXiv preprint arXiv:2303.02259}, 2023.

\bibitem{pei2010coordinated}
Yuanteng Pei, Matt~W Mutka, and Ning Xi.
\newblock Coordinated multi-robot real-time exploration with connectivity and
  bandwidth awareness.
\newblock In {\em IEEE International Conference on Robotics and Automation
  (ICRA)}, pages 5460--5465, 2010.

\bibitem{pei2013connectivity}
Yuanteng Pei, Matt~W Mutka, and Ning Xi.
\newblock Connectivity and bandwidth-aware real-time exploration in mobile
  robot networks.
\newblock {\em Wireless Communications and Mobile Computing}, 13(9):847--863,
  2013.

\bibitem{petravcek2021large}
Pavel Petr{\'a}{\v{c}}ek, V{\'\i}t Kr{\'a}tk{\`y}, Mat{\v{e}}j Petrl{\'\i}k,
  Tom{\'a}{\v{s}} B{\'a}{\v{c}}a, Radim Kratochv{\'\i}l, and Martin Saska.
\newblock Large-scale exploration of cave environments by unmanned aerial
  vehicles.
\newblock {\em IEEE Robotics and Automation Letters}, 6(4):7596--7603, 2021.

\bibitem{Podnar2006HumanTO}
Gregg Podnar, John~M. Dolan, Alberto Elfes, Marcel Bergerman, H.~Benjamin
  Brown, and Alan~D. Guisewite.
\newblock Human telesupervision of a fleet of autonomous robots for safe and
  efficient space exploration.
\newblock In {\em IEEE/ACM International Conference on Human-Robot
  Interaction}, 2006.

\bibitem{reardon2019communicating}
Christopher Reardon, Kevin Lee, John~G Rogers, and Jonathan Fink.
\newblock Communicating via augmented reality for human-robot teaming in field
  environments.
\newblock In {\em 2019 IEEE International Symposium on Safety, Security, and
  Rescue Robotics (SSRR)}, pages 94--101. IEEE, 2019.

\bibitem{rooker2007multi}
Martijn~N Rooker and Andreas Birk.
\newblock Multi-robot exploration under the constraints of wireless networking.
\newblock {\em Control Engineering Practice}, 15(4):435--445, 2007.

\bibitem{saboia2022achord}
Maira Saboia, Lillian Clark, Vivek Thangavelu, Jeffrey~A Edlund, Kyohei Otsu,
  Gustavo~J Correa, Vivek~Shankar Varadharajan, Angel Santamaria-Navarro,
  Thomas Touma, Amanda Bouman, et~al.
\newblock Achord: Communication-aware multi-robot coordination with
  intermittent connectivity.
\newblock {\em IEEE Robotics and Automation Letters}, 7(4):10184--10191, 2022.

\bibitem{schack2024sound}
Matthew~A Schack, John~G Rogers, and Neil~T Dantam.
\newblock The sound of silence: Exploiting information from the lack of
  communication.
\newblock {\em IEEE Robotics and Automation Letters}, 2024.

\bibitem{tian2024ihero}
Zhuoli Tian, Yuyang Zhang, Jinsheng Wei, and Meng Guo.
\newblock ihero: Interactive human-oriented exploration and supervision under
  scarce communication.
\newblock In {\em Robotics: Science and systems (RSS), Delft, Netherlands}.
  IEEE, 2024.

\bibitem{vaquero2018approach}
Tiago Vaquero, Martina Troesch, and Steve Chien.
\newblock An approach for autonomous multi-rover collaboration for mars cave
  exploration: Preliminary results.
\newblock In {\em International Symposium on Artificial Intelligence, Robotics,
  and Automation in Space (SAIRAS)}, 2018.

\bibitem{yamauchi1997frontier}
Brian Yamauchi.
\newblock A frontier-based approach for autonomous exploration.
\newblock In {\em IEEE International Symposium on Computational Intelligence in
  Robotics and Automation (CIRA)}, pages 146--151, 1997.

\bibitem{yamauchi1999decentralized}
Brian Yamauchi.
\newblock Decentralized coordination for multirobot exploration.
\newblock {\em Robotics and Autonomous Systems}, 29(2-3):111--118, 1999.

\bibitem{zavlanos2011graph}
Michael~M Zavlanos, Magnus~B Egerstedt, and George~J Pappas.
\newblock Graph-theoretic connectivity control of mobile robot networks.
\newblock {\em Proceedings of the IEEE}, 99(9):1525--1540, 2011.

\bibitem{zhou2023racer}
Boyu Zhou, Hao Xu, and Shaojie Shen.
\newblock Racer: Rapid collaborative exploration with a decentralized multi-uav
  system.
\newblock {\em IEEE Transactions on Robotics}, 2023.

\end{thebibliography}

\end{document}